\documentclass[doublespacing]{elsart}
\usepackage{macro_elsart}
\usepackage{epic,rotating}
\usepackage{epsfig,subfigure}
\usepackage{add_re_def}
\usepackage{latexsym,amsxtra,amssymb}
\usepackage{color}
\usepackage{float}

\begin{document}

\begin{frontmatter}
  \runauthor{Luc Brun and Walter Kropatsch}

  \title{Contains and Inside relationships within Combinatorial Pyramids}
  
  \author[greyc]{Luc Brun}, 
  \author[prip]{Walter Kropatsch\thanksref{asf}}
  
  \thanks[asf]{WK was supported by the Austrian Science Foundation
    under grants P14662-INF and FSP-S9103-N04}

\address[greyc]{
  GreYC -CNRS UMR 6072\\
  ENSICAEN\\
  6 Boulevard du Mar{\'e}chal Juin \\
  14045 Caen(France),\\
email: luc.brun@greyc.ensicaen.fr, fax: 33.231.45.26.98}

\address[prip]{
  Institute for Computer-Aided Automation\\
  Pattern Recognition and Image Processing Group\\
  Vienna Univ. of Technology- Austria }


\begin{abstract}

  Irregular pyramids are made of a stack of successively reduced
  graphs embedded in the plane. Such pyramids are used within the
  segmentation framework to encode a hierarchy of partitions. The
  different graph models used within the irregular pyramid framework
  encode different types of relationships between regions. This paper
  compares different graph models used within the irregular pyramid
  framework according to a set of relationships between regions. We
  also define a new algorithm based on a pyramid of combinatorial
  maps which allows to determine if one region contains the other
  using only local calculus.

\end{abstract}
\end{frontmatter}

\section{Introduction}
\label{sec:intro}

Graphs play an important role in computer vision and pattern
recognition since the birth of these fields. Graphs are used along the
overall process from the stimuli to the interpretation task:
hierarchical and non-hierarchical data structures for image
segmentation, graph matching for pattern recognition, graph clustering
for structural classification, and computation of a median
graph~\cite{jiang-01} for learning the structural properties of models.

Graphs are thus used both for low level image processing and high
level tasks. Different type of graphs being used for different types
of applications. However, in many computer vision tasks, the low image
segmentation stage cannot be readily separated from higher level
processing.  On the contrary, the segmentation algorithms should often
extract fine information about the partition in order to guide the
segmentation process according to the high level goal. This
information may be used to compare isolated regions or some local
configuration of regions to a model. There is thus a need to design
graph models for image segmentation which can be both efficiently
updated and allow to extract fine information about the partition.

\subsection{Relating Regions}
\label{subsec:relating}

\begin{figure}[htbp]
  \centering
\subfigure[]{ \epsfig{file=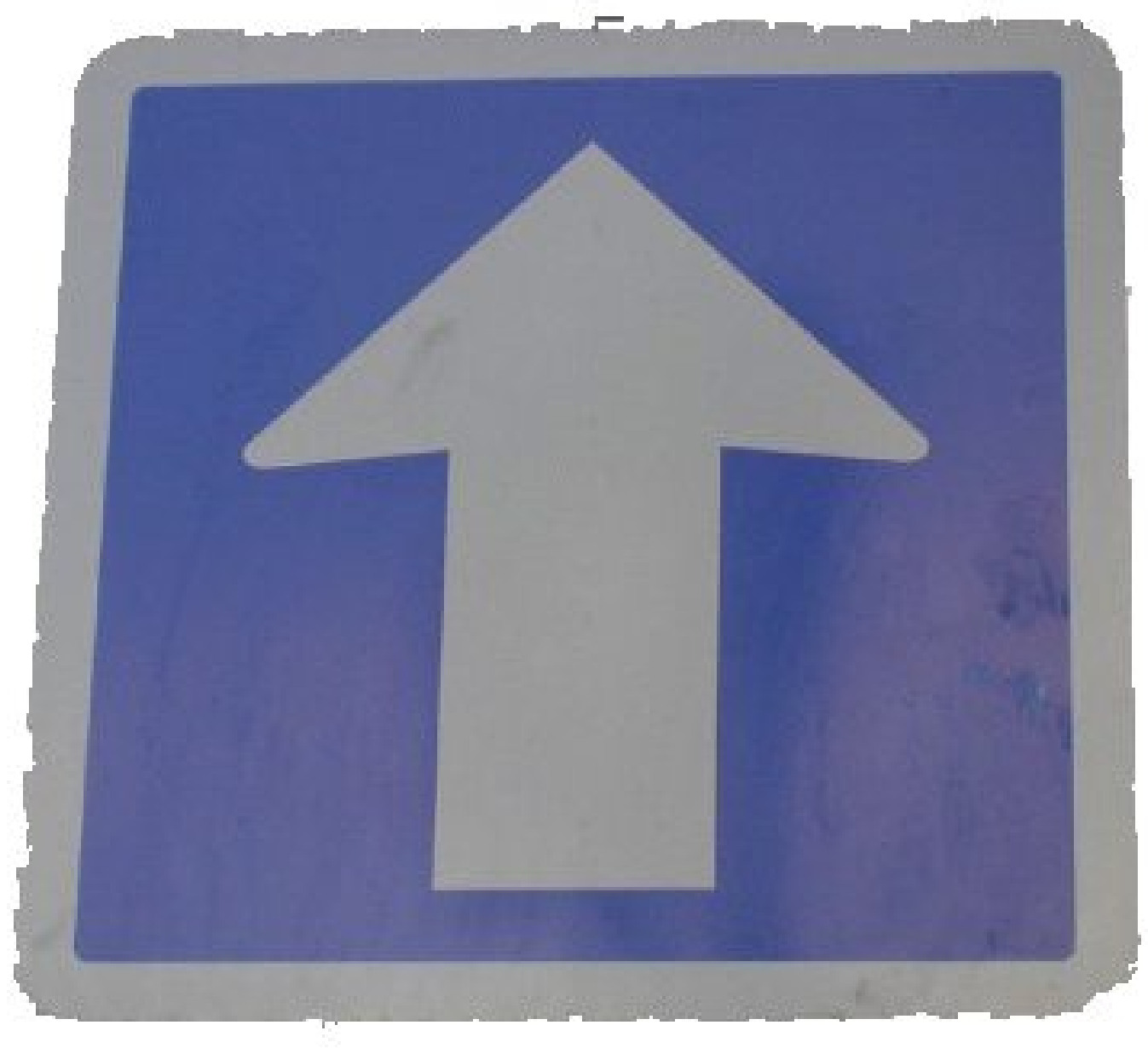,width=2cm}}
\subfigure[]{ \epsfig{file=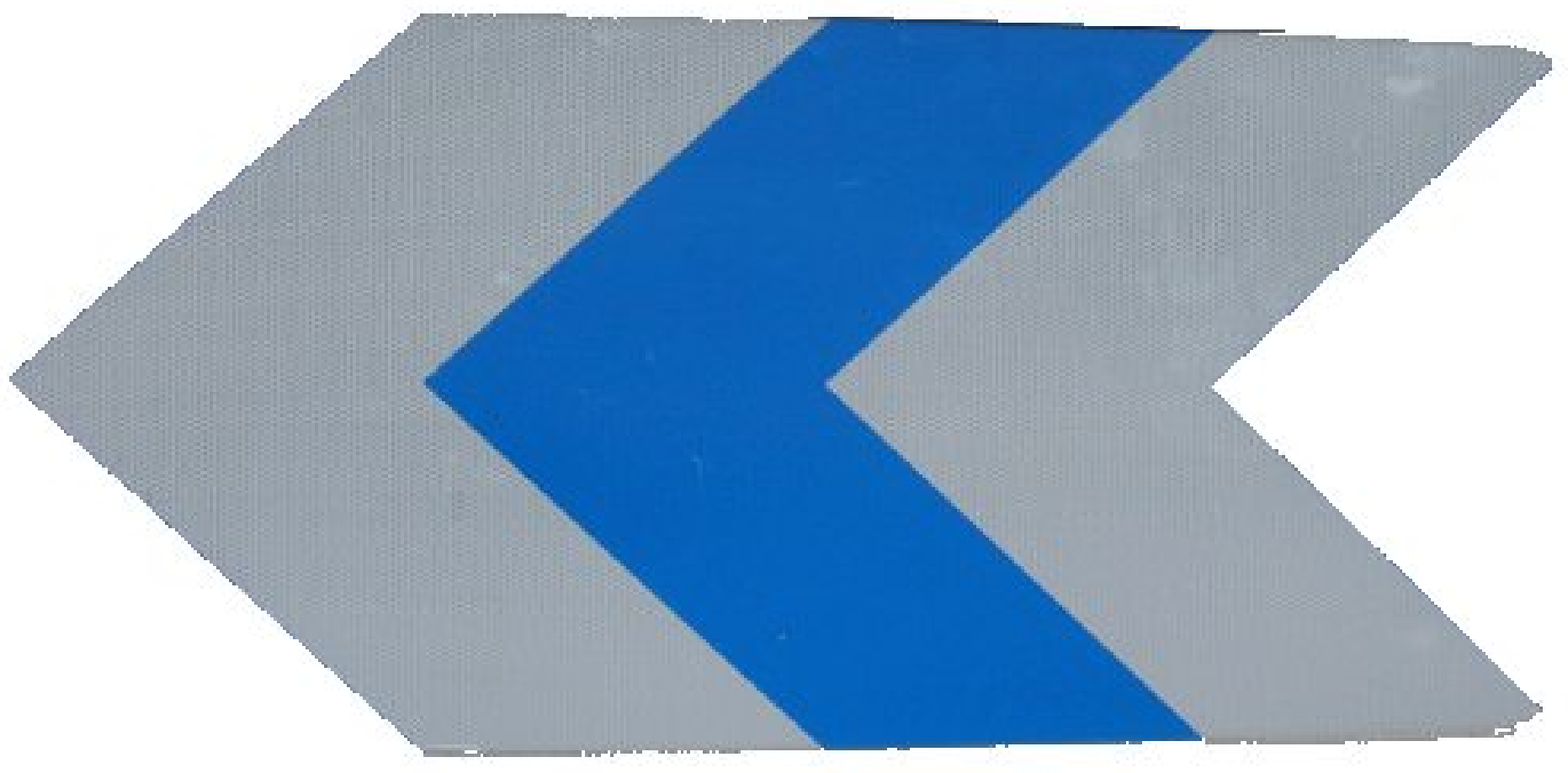,width=2cm}}
\subfigure[RAG(a)=RAG(b)]{ \epsfig{file=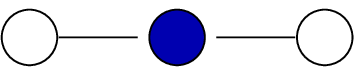,width=4.2cm}}
  \caption{The ideal segmentation of the two roadsigns (a) and  (b) are encoded by the same RAG (c).}
  \label{fig:fleche_flag}
\end{figure}

Different formalisms such as the RCC-8 defined by
Randel~\cite{randell-92} or the relationships defined by K. Shearer et
al.~\cite{shearer-01} in the context of graph matching may be used to
relate the regions of a partition. Within the particular context of
image segmentation, the following relationships may be defined from
these two models:\emph{meets}, \emph{contains}, \emph{inside}:
\begin{enumerate}

\item The \emph{meets} relationship means that two regions share at
least one common boundary. The different models used to encode
partitions either encode the existence of this common boundary or
create one relationship for each boundary between two regions. We
denote these two types of encodings \emph{meets\_exists} and
\emph{meets\_each}. The ability of the models to retrieve efficiently a
given common boundary between two regions is also an important feature
of these models.

\item The relationship \emph{A contains B} expresses the fact that
region B is inside region A. For example, the background of the road
sign in Fig.~\ref{fig:fleche_flag}(a) contains the upside arrow.

\item The \emph{inside} relationship is the inverse of the
\emph{contains} relation: A region \emph{B inside A} is contained in
A.

\end{enumerate}

 One additional relationship not directly handled by the models of
 Shearer and Randel may be defined within the hierarchical
 segmentation scheme. Indeed within such a framework a region defined
 at a given level of a hierarchy is \emph{composed of} regions defined
 at levels below.

The following relationships may thus be deduced from the relationships
defined by Shearer and Randel: The \emph{meets\_exists, meets\_each,
contains, inside} and \emph{composed of}.  Note that unlike meets
relationships, the contains and inside relations are asymmetric.  A
contains or inside relation between two regions allows thus to
characterize each of the regions sharing this relation.

\subsection{Region Adjacency Graph}
\label{subsec:rag}

 One of the most common graph data structure, within the segmentation
 framework is the Region Adjacency Graph (RAG). A RAG is defined from
 a partition by associating one vertex to each region and by creating
 an edge between two vertices if the associated regions share a common
 boundary. A RAG corresponds thus to a simple graph without any double
 edge between vertices nor self-loop.  Within a non-hierarchical
 segmentation scheme the RAG model is usually applied as a merging
 step to overcome the over-segmentation produced by the previous
 splitting algorithm~\cite{tweed-03}. Indeed, the existence of an edge
 between two vertices denotes the existence of at least one common
 boundary segment between the two associated regions which may thus be
 merged by removing this segment. Within this framework, the edge
 information may thus be interpreted as a possibility to merge the two
 regions identified by the vertices incident to the edge. Such a merge
 operation implies to collapse the two vertices incident to the edge
 into one vertex and to remove this edge together with any double edge
 between the newly created vertex and the remaining vertices.

The RAG model encodes thus only the existence of a common edge between
two regions (the \emph{meets\_exists} relationship). Moreover, the
existence of a common edge between two vertices does not provide
enough information to differentiate a meets relationship from a
\emph{contains} or \emph{inside} one. This drawback is illustrated on
an ideal segmentation of two roadsigns (Fig.~\ref{fig:fleche_flag})
which are encoded by a same RAG. The road sign (a) defines two nested
\emph{contains} relationships. Indeed, the white border
\emph{contains} the background which \emph{contains} itself the
symbol. On the other hand the road sign (b) corresponds to two meets
relationships between the central region and its two white neighbors.

\subsection{Combinatorial Maps}
\label{subsec:combi_maps}

A 2D combinatorial map may be understood as a planar graph encoding
explicitly the orientation in the plane. Each connected component of a
partition (a connected set of regions) may be encoded by a 2D
combinatorial map up to an homeomorphism~\cite{braquelaire-96-1}. One
of the main insight of such models compared to a RAG data structure is
their ability to be efficiently updated after both split and merge
operations.

The combinatorial map formalism allows to encode each connected
boundary between two regions by one edge. The models based on
combinatorial maps encode thus the \emph{meets\_each}
relationship. However, within the combinatorial map framework two
connected components of a partition will be encoded by two
combinatorial maps without any information about the respective
positioning of the two components. The models based on combinatorial
maps have thus designed additional data structure like the inclusion
tree~\cite{fiorio-95} or the Parent-Child
relationships~\cite{braquelaire-96-1,brun-02-5} to encode the
\emph{contains} and \emph{inside} relationships. Using these models
any modification of the partition implies to update both the
combinatorial maps and the additional data structures.


\subsection{Segmentation Hierarchies}
\label{subsec:irr_pyr}

Data structures used within the hierarchical segmentation framework
encode a stack of partitions successively simplified by region
merging. Irregular pyramids models introduced by Meer and
Montanvert~\cite{montanvert-91} encode each partition by a
graph where each vertex is associated to one region.  At each level of
the pyramid a region is obtained by the merge of a connected set of
regions at the level below. The resulting region is called the parent
of the merged regions. These last regions correspond to the child of
the region at the level above.  The models based on the irregular
pyramid framework encode thus naturally the \emph{composed of}
relationship.  In order to preserve the efficiency of a hierarchical
data structure, the size of models encoding each partition of the
hierarchy must be strictly decreasing according to the level.  This
last constraint forbids the use of an additional data structure
similar to the structures used for combinatorial maps models in order
to store \emph{contains} and \emph{inside} relationships. Indeed,
\emph{contains} and \emph{inside} relationships between regions may
both be removed and created along the different levels of the
pyramid. The use of such an additional data structure may thus violate
the strictly decreasing size of the models according to the levels.

\subsection{Overview}
\label{subsec:overview}

The aim of this paper is twofold: We firstly provide an introduction
to the main data structures used within the hierarchical segmentation
framework according to the set of relationships previously defined
(Section~\ref{subsec:relating}). Secondly, we present an efficient
computation of the \emph{contains} and \emph{inside} relationships
within the irregular pyramid framework.  The remaining of this paper
is thus organized as follows: Section~\ref{sec:irr_pyr} presents two
models belonging to the irregular pyramid framework together with
their properties relative to the relationships previously defined.
Section~\ref{sec:combi_maps} describes the combinatorial map model and
its main properties.  Section~\ref{sec:combi_pyr} describes the
construction scheme and the main properties of a pyramid of
combinatorial maps : A combinatorial Pyramid. Finally,
Section~\ref{sec:inclusion} presents one algorithm computing the
contains and inside information using only local calculus.


\section{Simple and Dual Graph Pyramids}
\label{sec:irr_pyr}
The irregular pyramids are defined as a stack of successively reduced
graphs, each graph being built from the graph below by selecting a set
of vertices named surviving vertices and mapping each non-surviving
vertex to a surviving vertex~\cite{meer-89,montanvert-91}. Using such a
framework, the graph $G_{l+1}=(V_{l+1},E_{l+1})$ defined at level
$l+1$ is deduced from the graph defined at level $l$ by the following
steps:
\begin{enumerate}
\item Select the vertices of $G_{l+1}$ among $V_{l}$. These vertices
are the surviving vertices of the decimation process, $V_{l+1}\subset V_l$.
\item Each non-surviving vertex connects to a surviving vertex by one
edge of $G_l$. The set of vertices attached to each surviving vertex
defines a partition of $V_l$.

\item Define the adjacency relationships between the vertices of
$G_{l+1}$ in order to define $E_{l+1}$.
\end{enumerate}

\subsection{Simple graph Pyramids}
\label{subsec:simple_graph}

In order to obtain a decimation ratio greater than 1 between two
successive levels, Meer~\cite{meer-89} imposes the following
constraints on the set of surviving vertices:
\begin{eqnarray}
  \forall v\in V_l-V_{l+1}\; \exists v'\in V_{l+1}:\; (v,v')\in
  E_l\label{eq-mis1}\\ 
  \forall (v,v')\in V_{l+1}^2:\; (v,v')\not\in  E_l\label{eq-mis2}
\end{eqnarray}
Constraint~(\ref{eq-mis1}) insures that each non-surviving vertex is
adjacent to at least a surviving vertex. Constraint~(\ref{eq-mis2})
insures that two adjacent vertices cannot both survive. These
constraints define a {\em maximal independent set}
(MIS)~\cite{meer-89,montanvert-91}.

Given the set of surviving vertices, different
methods~\cite{montanvert-91,brun-02-9} may be used to link each
non-surviving vertex to one of its surviving neighbor. For example,
Montanvert~\cite{montanvert-91} attaches each non-surviving vertex to
its closest surviving neighbor according to a difference between the
outcome of a random variable attached to each vertex. The set of
non-surviving vertices connected to a surviving vertex defines its
reduction window and thus the parent child relationship between two
consecutive levels.

The final set of surviving vertices defined on $V_l$ corresponds to
the set of vertices $V_{l+1}$ of the reduced graph
$G_{l+1}=(V_{l+1},E_{l+1})$. The set of edges $E_{l+1}$ of $G_{l+1}$
is defined by connecting by an edge in $G_{l+1}$ any couple of
surviving vertices having adjacent children.

Two surviving vertices are thus connected in $G_{l+1}$ if they are
connected in $G_l$ by a path of length lower or equal than $3$.  Two
reduction windows adjacent by more than one path of length lower or
equal than $3$ will thus be connected by a single edge in the reduced
graph. The stack of graphs produced by the above decimation process is
thus a stack of simple graphs each simple graph encoding only the
existence of one common boundary between two regions (the
\emph{meeets\_exists} relationship).  Moreover, as mentioned in
Section~\ref{subsec:rag} the RAG model which corresponds to a simple
graph does not allow to encode \emph{contains} and \emph{inside}
relationships.

\subsection{Construction of Dual Graph Pyramids}
\label{subsec:dual_graph}

 The dual graph pyramids introduced by Willersinn and
 Kropatsch~\cite{willersinn-94} use an alternative construction
 scheme. Within the dual graph pyramid framework the reduction process
 is performed by a set of edge contractions. The edge contraction
 collapses two adjacent vertices into one vertex and removes the edge.
 Many edges except self-loops can be contracted independently of each
 other and also in parallel. In order to avoid contracting a self-loop
 these edges should not form cycles, e.g. form a forest. This set of
 edges is called a contraction kernel.

 The contraction of a graph reduces the number of vertices while
 maintaining the connections to other vertices. As a consequence some
 redundant edges may occur. These edges belong to one of the following
 categories:
 \begin{itemize}
 \item Redundant double edge: These edges encode multiple adjacency
 relationships between two vertices and define degree two faces.  They
 can thus be characterized in the dual graph as degree two dual
 vertices.  In terms of partition's encoding these edges correspond to
 an artificial split of one boundary between two regions.
   
 \item Empty self-loop: These edges correspond to a self-loop with an
 empty inside. These edges define thus degree one faces and are
 characterized in the dual graph as degree one vertices. Such edges
 encode artificial inner boundaries of regions.
 \end{itemize}
 
 Both double edges and empty self-loops do not encode relevant
 topological relations and can be removed without any harm to the
 involved topology~\cite{willersinn-94}. The removal of such edges is
 called a dual decimation step and the set of removed edges is called
 a removal kernel. Such a kernel defines a forest of the dual graph.
 
\subsection{Dual Graph Pyramids and multiple boundaries}
\label{subsec:dual_graph_meet_each}
 Given one tree of a contraction kernel, the contraction of its edges
 collapses all the vertices of the tree into a single vertex and keeps
 all the connections between the vertices of the tree and the
 remaining vertices of the graph. The multiple boundaries between the
 newly created vertex and the remaining vertices of the graph are thus
 preserved. Each graph of a dual graph pyramid encodes thus the
 \emph{meets\_each} relationships. This property is not modified by the
 application of a removal kernel which only removes redundant edges.
 
\subsection{Dual Graph Pyramids and the inside relationship}
\label{subsec:dual_graph_inside}

\begin{figure}[htbp]
   \centering
   \epsfig{file=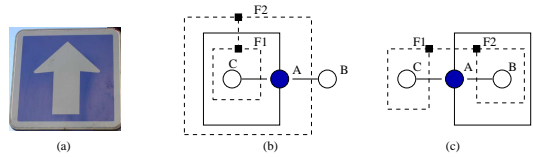,height=2cm}
   \caption{The Graph (b) defines the top of a dual graph pyramid encoding an ideal segmentation of (a). The self loop incident to vertex A may surround either vertex B or C without changing the incidence relations between vertices and faces. The dual vertices associated to faces are represented by filled boxes ($\blacksquare$). Dual edges are represented by dashed lines.}
   \label{fig:fleche_inside}
 \end{figure}

 Due to the forest requirement, the encoding of the adjacency between
 two regions one inside the other will be encoded by two edges
 (Fig.~\ref{fig:fleche_inside}): One edge encoding the common border
 between the two regions and one self-loop incident to the vertex
 encoding the surrounding region.  One may think to characterize the
 inside relationship by the fact that the vertex associated to the
 inside region should be surrounded by the self-loop. However, as
 shown by Fig.~\ref{fig:fleche_inside}(c) one may exchange the
 surrounded vertex without modifying the incidence relationships
 between both vertices and faces. Two dual graphs being defined by
 these incidence relationships one can exchange the surrounded vertex
 without modifying the encoding of the graphs. This last remark shows
 that the \emph{inside/contains} relationships cannot be characterized
 locally within the dual graph framework.


\section{Combinatorial maps}
\label{sec:combi_maps}

Combinatorial maps and generalized combinatorial maps define a general
framework which allows to encode any subdivision of nD topological
spaces orientable or non-orientable with or without boundaries.
Recent trends in combinatorial maps apply this framework to the
segmentation of 3D images~\cite{braquelaire-01,bertrand-01} and the
encoding of 2D~\cite{brun-03-1,brun-05} and nD~\cite{damiand-05}
hierarchies.

The remaining of this paper will be based on 2D combinatorial maps
which will be just called combinatorial maps. A combinatorial map may
be deduced from a planar graph by splitting each edge into two half
edges called darts. An edge connecting two vertices is thus composed
of two darts, each dart belonging to only one vertex. The relation
between two darts $d_1$ and $d_2$ associated to the same edge is
encoded by the permutation $\alpha$ which maps $d_1$ to $d_2$ and $d_2$ to
$d_1$. The permutation $\alpha$ is thus an involution and its
cycles\footnote{the cycle of a dart $d$ associated to a permutation
$\pi$ on the set of darts is the sequence $(d,\pi(d),\pi^2(d),\dots,\pi^n(d))$
with $\pi^n(d)=d$. Since the set of darts is finite $n$ is defined for
any dart and any permutation $\pi$. The $\pi$ orbit of a dart $d$
correponds to the same set of darts as its cycle but without any
ordering between darts.} are denoted by $\alpha^*(d)$ for a given dart
$d$. These cycles encode the edges of the graph.  The sequence of
darts encountered when turning around a vertex is encoded by the
permutation $\sigma$. Using a counter-clockwise orientation, the cycle
$\sigma^*(d) $ encodes the set of darts encountered when turning
counter-clockwise around the vertex encoded by the dart $d$. A
combinatorial map can thus be formally defined by
$G=(\dartset{},\sigma,\alpha)$, where $\dartset{}$ is the set of darts and $\sigma$,
$\alpha$ are two permutations defined on $\dartset{}$ such that $\alpha$ is an
involution.\footnote{$\pi$ is an involution on $\dartset$ if $\pi\circ\pi(d)=d$
for any dart $d$ in $\dartset$}

Given a combinatorial map $G=(\dartset,\sigma,\alpha)$, its dual is
defined by $\dual{G}=(\dartset,\varphi,\alpha)$ with
$\varphi=\sigma\circ\alpha$.  The cycles of the permutation $\varphi$
encode the set of darts encountered when turning around a face of $G$.

We can state one of the major difference between a combinatorial map
and an usual graph encoding of a partition. Indeed, a combinatorial
map may be seen as a planar graph with a set of vertices (the cycles
of $\sigma$) connected by edges (the cycles of $\alpha$).  However, compared to
an usual graph encoding a combinatorial map encodes additionally the
local orientation of edges around each vertex thanks to the order
defined within each cycle of $\sigma$.

\begin{figure}[htbp]
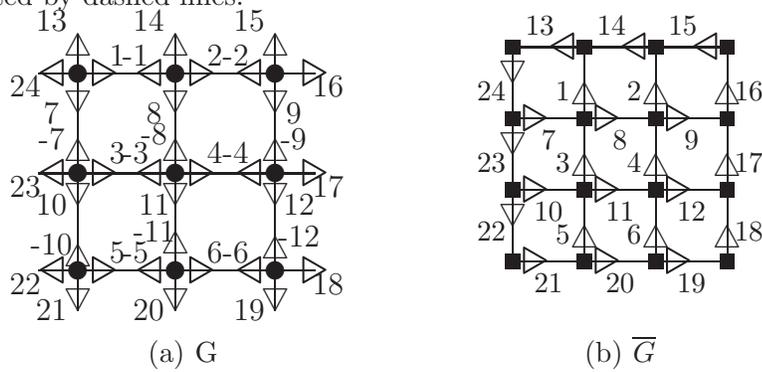

  \input{nodes.pic} \input{brins.tex}
\begin{center}
  \unitlength 0.3mm 
\mbox{ }\hfill
\subfigure[G]{
  \unitlength 0.35mm 
\begin{picture}(100.00,90.00) 
  \input{grille.pin}
\end{picture}
}\hfill
\subfigure[\dual{G}]{
  \unitlength 0.38mm 
\begin{picture}(100.00,92.00)
  \input{grille_dual.pin}
\end{picture}
}
\hfill\mbox{ }
\end{center}
\caption{A $3\times 3$ grid encoded by a combinatorial map}
\label{fig:grid}
\end{figure}

Fig.~\ref{fig:grid} illustrates the encoding of a $3\times3$ $4$-connected
discrete grid by a combinatorial map $G$.  The involution $\alpha$ is
implicitly encoded by the sign in Fig.~\ref{fig:grid}(a) and (b). We
have thus $\alpha(d)=-d$ for all darts on these figures.

Since $G$ encodes a planar sampling grid, each vertex of $\dual{G}$
(Fig.~\ref{fig:grid}(b)) is associated to a corner of a pixel.  For
example, the top left pixel of the $3\times 3$ grid is encoded by the $\sigma$
cycle $\sigma^*(1)=(1,13,24,7)$ (top left vertex and square in
Fig.~\ref{fig:grid}(a) and (b)).  The top-left corner of this pixel is
encoded by the $\varphi$ cycle: $\varphi^*(24)=(24,-13)$ (top left dual vertex of
Fig.~\ref{fig:grid}(b)).  Moreover, each dart may be understood in
this combinatorial map as an oriented crack, i.e. as a side of a pixel
with an orientation.  For example, the dart $1$ in
Fig.~\ref{fig:grid}(b) encodes the right side of the upper-left pixel
oriented from bottom to top. The $\varphi$, $\alpha$ and $\sigma$ cycles of a dart may
thus be respectively understood as elements of dimensions $0$, $1$ and
$2$.
  
Each dart of a combinatorial map $G$, encoding a planar sampling grid
may thus be interpreted as an oriented crack and associated to a point
encoding the coordinates of a pixel's corner and one move encoding the
orientation on the crack associated to the dart.


\section{Embedding and Orientation}
\label{sec:combi_pyr}

As in the dual graph pyramid scheme~\cite{Kropatsch95b}
(Section~\ref{sec:irr_pyr}) a combinatorial pyramid is defined by an
initial combinatorial map successively reduced by a sequence of
contraction or removal operations. The initial combinatorial map
encodes a planar sampling grid (Section~\ref{sec:combi_maps})
or a first segmentation and the remaining combinatorial maps of a
combinatorial pyramid encode a stack of image partitions successively
reduced. Such combinatorial maps are thus embedded
(Section~\ref{subsec:dart_embed}). As mentioned in
Section~\ref{sec:combi_maps} a combinatorial map may be understood as
a dual graph with an explicit encoding of the orientation of the edges
incident to each vertex. This explicit encoding of the orientation is
preserved within the combinatorial pyramid using contraction and
removal operations equivalent to the operations used for dual graphs
but which preserve the orientations of edges around the vertices of
the reduced combinatorial maps~\cite{brun-03-1,brun-05}.

Contraction operations are controlled by contraction kernels (CK).
The removal of redundant edges is performed as in the dual graph
reduction scheme by a removal kernel. This kernel is however
decomposed in two sub-kernels : A removal kernel of empty self-loops
(RKESL) which contains all darts incident to a degree 1 dual vertex
and a removal kernel of empty double edges (RKEDE) which contains all
darts incident to a degree 2 dual vertex.  These two removal kernels
are defined as follows : The removal kernel of empty self-loops RKESL
is initialized by all self-loops surrounding a dual vertex of degree
1. RKESL is further expanded by all self-loops that contain only other
self-loops already in RKESL until no further expansion is
possible. For the removal of empty double edges RKEDE we ignore all
empty self-loops in RKESL in computing the degree of the dual
vertex. Note that the successive application of a RKESL and a RKEDE is
equivalent to the application of a removal kernel defined within the
dual graph framework. Both contraction and removal operations defined
within the combinatorial pyramid framework are thus defined as is the
dual graph framework but additionally preserve the orientation of edges
around each vertex.  Further details about the construction scheme of
a Combinatorial Pyramid may be found in~\cite{brun-03-1}.

\subsection{What is inside ?}
\label{subsec:problem}
Combinatorial pyramids are thus built using the same framework as dual
graphs pyramids. The use of a contraction kernel within the
construction scheme of a combinatorial pyramid allows to encode
multiple adjacency between regions thanks to multiple edges between
their associated vertices. Therefore, as in the dual graph framework,
combinatorial pyramids preserve the \emph{meets\_each} relationship
(Section~\ref{subsec:dual_graph_meet_each}). Note that the explicit
encoding of the orientation within the combinatorial pyramid framework
does not interfere with this last property.

\begin{figure}[htbp]
  \centering
     \epsfig{file=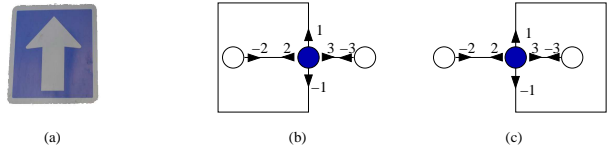,height=2cm}
  \caption{The encoding of an ideal segmentation of a road sign (a) by the top level combinatorial map of a pyramid may be drawn using either (b) or (c).}
  \label{fig:combi_inside}
\end{figure}

Moreover, as in the dual graph framework
(Section~\ref{subsec:dual_graph_inside}), an inside relationship
between two regions is encoded by two edges: one edge encodes the
common border between the two regions while the other encodes a
self-loop incident to the vertex associated to the surrounding
region. Let us consider the example already used for dual graph
pyramids (Section~\ref{subsec:dual_graph_inside},
Fig.~\ref{fig:fleche_inside}).  Fig.~\ref{fig:combi_inside} shows the
encoding of the ideal segmentation of the road sign using
combinatorial maps. As shown by Fig.~\ref{fig:combi_inside} (b) and
(c), one can exchange the surrounded vertex without changing the order
of the darts around the vertex $\sigma^*(1)$.  Therefore, the two drawings
shown in Fig.~\ref{fig:combi_inside}(b) and (c) are encoded by the
same combinatorial map. One cannot determine from the formally
specified combinatorial map which part is inside and which is
contained. This ambiguity may also be characterized using the cycle
$\sigma^*(1)$ of the vertex incident to the self-loop. Indeed, this cycle
is equal to $\sigma^*(1)=(1,2,-1,3)$.  Since $(1,-1)$ is a self-loop the
neighbors of $\sigma^*(1)$ have in their $\sigma$ cycles the $\alpha$ successors of
the two darts $2$ and $3$. The ambiguity about the drawing of the
self-loop is characterized on the cycle $\sigma^*(1)$ by the fact that we
can not deduce from this cycle if the dart $2$ is between $1$ and $-1$
or if on the contrary $3$ is between $-1$ and $1$. This ambiguity
arises thus because the two darts $1$ and $-1$ play a symmetric role
in $\sigma^*(1)$. We can thus state the two following points from the above
discussion:
\begin{enumerate}
\item Combinatorial pyramids preserve the \emph{meets\_each}
relationship.

\item An \emph{inside} relationship 'A inside B' is always associated with a
self-loop incident to B. However, a non-redundant self-loop at B does
not always identify the inside region.
\end{enumerate}

\subsection{Implicit encoding  of a combinatorial pyramid}
\label{subsec:const_scheme}





Let us consider an initial combinatorial map $G_0=(\dartset,\sigma,\alpha)$ and
a sequence of kernels $K_1,\dots,K_n$ successively applied on $G_0$ to
build the pyramid. Each combinatorial map $G_i=(\survive_i,\sigma_i,\alpha_i)$
is defined from $G_{i-1}=(\survive_{i-1},\sigma_{i-1},\alpha_{i-1})$ by the
application of the kernel $K_{i}$ on $G_{i-1}$ and the set of darts
$\survive_{i}$ is equal to $\survive_{i-1}\setminus K_i$. We have thus:
\begin{equation}
  \label{eq:incl_set_darts}
  \survive_{n+1}\subset\survive_n\subset \dots\survive_1\subset\dartset 
\end{equation}
The set of darts of each reduced combinatorial map of a pyramid is
thus included in the base level combinatorial map. This last property
allows us to define the two following functions:
\begin{enumerate}
\item one function $state$ from $\{1,\dots,n\}$ to the states
  $\{CK, RKESL, RKEDE\}$ which specifies the
  type of each kernel.

\item One function $level$ defined for all darts in $\dartset$ such
that $level(d)$ is equal to the maximal level where $d$ survives:
\[
\forall d\in \dartset\;level(d)=
\begin{array}[t]{r}
Max\{i\in\{1,\dots,n+1\} \,|\,
d\in\survive_{i-1}\} \\
\end{array}
\]
a dart $d$ surviving up to the top level has thus a level equal to
$n+1$. Note that if $d\in K_i, i\in \{1,\dots,n\}$ then $level(d)=i$.

\end{enumerate}
We have shown~\cite{brun-02-9,brun-03-1} that the sequence of reduced
combinatorial maps $G_0,\dots,G_{n+1}$ may be encoded without any loss
of information using only the base level combinatorial map $G_0$ and
the two functions $level$ and $state$. Such an encoding is called an
\emph{implicit encoding} of the pyramid.

The receptive field of a dart $d\in \survive_i$ corresponds to the set
of darts reduced to $d$ at level $i$~\cite{brun-03-1,brun-02-9}. Using
the implicit encoding of a combinatorial pyramid, the receptive field
$RF_i(d)$ of $d\in \survive_i$ is defined as a sequence $d_1.\dots.d_q$
of darts in $\dartset$ by $d_1=d$, $d_2=\sigma_0(d)$ and for each $j$
in  $\{3,\dots,q\}$ :
\begin{equation}
  d_j = \left\{
    \begin{array}{ll}
         \varphi_0(d_{j-1}) & \mbox{if }state(level(d_{j-1}))=CK\\
         \sigma_0(d_{j-1}) &  \mbox{if }state(level(d_{j-1}))\in \{RKEDE,RKESL\}\\
       \end{array}
     \right.
   \label{eq:trav_rf}
 \end{equation}
The dart $d_q$ is defined as the last dart of the sequence which have
been contracted or removed below the level $i$. Therefore, the
successor of $d_q$ according to equation~\ref{eq:trav_rf}, $d_{q+1}$
satisfies $level(d_{q+1})>i$. Moreover, we have
shown~\cite{brun-03-1,brun-02-9} that $d$, $d_q$ and $d_{q+1}$ are
additionally connected by the two following relationships:
\begin{equation}
      \sigma_i(d)=d_{q+1} \mbox{ and }      \alpha_i(d)=\alpha_0(d_q)
  \label{eq:rf_succ}
\end{equation}
Note that these two last relationships allow to retrieve any reduced
combinatorial map of the pyramid from its base.

The implicit encoding of combinatorial pyramids is thus based on the
fact that the set of darts of any reduced combinatorial map is
included in the initial combinatorial map
(equation~\ref{eq:incl_set_darts}).  The two functions $state$ and
$level$ which are based on this property allow to encode the whole
sequence of reduced combinatorial map without loss of
information~\cite{brun-03-1,brun-02-9}.

\subsection{Dart's embedding and  Segments}
\label{subsec:dart_embed}

In the RAG a region corresponds to a vertex and two regions are
connected by an edge if the two regions share a boundary. In the
Combinatorial Map, vertices and edges correspond to $\sigma$ and $\alpha$ cycles
respectively. Therefore, each dart $d\in \survive_i$ encodes a boundary
between the regions associated to $\sigma_i^*(d)$ and
$\sigma_i^*(\alpha_i(d))$. Moreover, in the lower levels of the pyramid the two
vertices of an edge may belong to a same region. We call the
corresponding boundary segment an \emph{internal boundary} in contrast
to an \emph{external boundary} which separates two different regions
of a RAG. The receptive field of $d$ at level $i$ ($RF_i(d)$) contains
both darts corresponding to this boundary and additional darts
corresponding to internal boundaries.  The sequence of external
boundary darts contained in $RF_i(d)$ is denoted by $\partial RF_i(d)$ and is
called a \emph{segment}. The order on $\partial RF_i(d)$ is deduced from the
receptive field $RF_i(d)$. Given a dart $d\in \survive_i$, the sequence
$\partial RF_i(d)=d_1,\dots,d_q$ is retrieved by~\cite{brun-02-9}:
\begin{equation}
d_1=d \mbox{ and }\forall j\in\{1,\dots,q-1\} \;d_{j+1}=\varphi_0^{n_j}(\alpha_0(d_j))\\
\label{eq:trav_border}
\end{equation}
The dart $d_q$ is the last dart of $\partial RF_i(d)$ which belongs to a
double edge kernel.  This dart is thus characterized using
equation~\ref{eq:rf_succ} by $d_q=\alpha_0(\alpha_i(d))$. Note that each dart of
the base level corresponds to an oriented crack
(Section~\ref{sec:combi_maps}). A segment corresponds thus to a
sequence of oriented cracks encoding a connected boundary between two
regions~\cite{brun-02-9}.

The value $n_j$ is defined for each $j\in \{1,\dots,q-1\}$ by :
\begin{equation}
\label{eq:trav_border_n}
n_j=Min\{ k\in \NN^*\;| \; state(level(\varphi_0^k(\alpha_0(d_j))))=RKEDE\}.
\end{equation}
A segment may thus be interpreted as a maximal sequence, according to
equation~\ref{eq:trav_border}, of darts removed as double edges. Such
a sequence connects two darts ($d$ and $\alpha_0(d_q)=\alpha_i(d)$) surviving up
to level $i$. The retrieval of the boundaries using
equations~\ref{eq:trav_border} and~\ref{eq:trav_border_n} is one of
the major reason which lead us to distinguish empty self-loop removal
kernels and double edges.

Let us additionally note that if $G_0$ encodes the $4$-connected
planar sampling grid, each $\varphi_0$ cycle is composed of at most $4$
darts (Fig.~\ref{fig:grid}(b)).  Therefore, the computation of
$d_{j+1}$ from $d_j$ (equation~\ref{eq:trav_border}) requires at most
$4$ iterations and the determination of the whole sequence of cracks
composing a boundary between two regions is performed in a time
proportional to the length of this boundary.

\subsection{Computing Segment's Orientation}
\label{subsec:seg_or}

As mentioned in Section~\ref{sec:combi_maps}, each oriented crack
associated to an initial dart may be encoded by the position of its
starting point and one move.  The move of an initial dart $d$ is
denoted by $move(d)$. If the initial combinatorial map $G_0$ encodes a
square grid, the move associated to each dart belongs to the set
$\{right,up,left, down\}$. These initial moves are encoded using
Freeman's codes: right,up, left and down are numbered from 0 to 3. The
angle between two moves $m_1$ and $m_2$ denoted by $(m_1,m_2)\sphat~$
is then defined as: $(m_1-m_2)~mod~4$ where $mod$ corresponds to the
operator modulo. This angle is thus equal to: $+1$ if the two oriented
cracks define a clockwise ${90^\circ}$ turns, -1 if the two oriented crack
define a counter-clockwise ${90^\circ}$ turns, 0 if the two oriented crack
correspond to a same move and $2$ if the two oriented cracks
correspond to opposite moves. Note that these angles may be associated
to the RULI code (Right turn, U turn, Left turn and Identical) defined
by Ferm{\"u}ller and Kropatsch~\cite{Fermueller94a}. Indeed, the angles
associated to the R, U,L and I codes are respectively equal to +1, 2,
-1 and 0. Since the sequences of moves considered in this work do
define U turn, we consider an angle of $2$ between two moves as
undefined.

Given a dart $d$ of $G_i$, let us denote respectively by $Fm(d)$ and
$Lm(d)$ the moves of the first and last oriented cracks of the segment
associated to $d$.  If $\partial RF_i(d)=d_1\dots d_q$ we have $d_1=d$ and
$d_q=\alpha_0(\alpha_i(d))$ (equation~\ref{eq:rf_succ}) and $Fm(d)=move(d_1)$,
$Lm(d)=move(d_q)$. The two darts $d_1$ and $d_q$ may thus be retrieved
in constant time from $d$. Moreover the moves of $d_1$ and $d_q$ are
retrieved using a correspondence between the oriented cracks and the
initial darts. This correspondence may be defined using any implicit
numbering of the initial darts (see e.g.  Fig.~\ref{fig:grid}(a)). The
values of $Fm(d)$ and $Lm(d)$ may thus be retrieved without additional
memory requirement and in constant time using an appropriate numbering
of the initial darts.

Given a dart $d$ in $G_i$, and the sequence of darts $d_1\dots d_q$ in
$G_0$ encoding its segment, the properties of the segments together
with the properties of the combinatorial pyramids~\cite{brun-02-9}
induce the two following properties:
\begin{eqnarray}
  \forall j\in \{1,\dots,p-1\}\quad move(d_j)^{-1}\neq move(d_{j+1})  \label{eq:move_in_seg}\\
  Lm(d)\neq Fm(\sigma_i(d))^{-1} \label{eq:move_between_seg}
\end{eqnarray}
where $move(d_j)$ denotes the move of the oriented crack associated to
$d_j$ and $move(d_j)^{-1}$ is the opposite of the move of $d_j$
(e.g. $right^{-1}=left$). 

Equation~\ref{eq:move_in_seg} states that two successive moves within
a segment cannot be opposite. This property is induced by the fact
that one segment cannot contain twice a same crack with two
orientations. Equation~\ref{eq:move_between_seg} states that the first
move of the $\sigma_i$ successor of a dart $d$ cannot be the opposite of
the last move of $d$. Otherwise, the dart $d$ would be an empty
self-loop of $G_i$ which is refused by hypothesis. 



Given the angle between two successive oriented cracks we define the
orientation of a dart as the sum of the angles between the oriented
cracks along its associated segment. Given a dart $d$ in $G_i$ the
orientation of $d$ is  defined by:

  \begin{equation}
    \label{eq:or_dart}
    or(d)=\sum_{j=1}^{q-1} \left(move(d_j),move(d_{j+1})\right)\sphat
  \end{equation}

where $d_1\dots d_q$ is the the sequence of initial darts encoding the
segment associated to $d$. Note that $(move(d_j),move(d_{j+1})\sphat$
cannot be undefined for any $j\in \{1,\dots,q-1\}$
(equation~\ref{eq:move_in_seg}).

The orientation of a dart may be computed on demand using
equation~\ref{eq:or_dart} or may be attached to each dart and updated
during the construction of the pyramid.  Indeed, let us consider two
successive double darts $d_1$ and $d_2$ at one level of the
pyramid. If $d_1$ survives at the above level its orientation may be
updated by~\cite{braquelaire-96-1,brun-02-5}:
\begin{equation}
  \label{eq:or_update}
  or(d_1)=or(d_1)+or(d_2)+(Lm(d_1),Fm(d_2))\sphat
\end{equation}
Note that this last formula may be extended to the removal of a
sequence  of successive double edges. 

The dart's orientation may thus be computed by fixing the orientation
of all initial darts to $0$ and updating the dart's orientation using
equation~\ref{eq:or_update} during the removal of each double edge
kernel.

Let us consider a sequence $d_1\dots d_p$ in $G_i$ such that
$d_{j+1}=\sigma_i(d_j)$ for all $j$ in $\{1,\dots,p-1\}$ and $d_p\neq
\alpha_i(d_1)$. We say that such a sequence defines a closed boundary if
$\alpha_i(d_p)$ and $d_1$ are incident to a same dual vertex
e.g. $d_1\in \varphi_i^*(\alpha_i(d_p))$. The orientation of such a sequence is
defined by:
\begin{equation}
\label{eq:or_seq}
or(d_1\dots d_p)=\sum_{j=1}^{p-1}\left(or(d_j)+(Lm(d_j),Fm(d_{j+1}))\sphat\right)+or(d_p)
\end{equation}
The quantity $(Lm(d_p),Fm(d_1))\sphat~$ has to be added to
$or(d_1\dots d_p)$ if the sequence defines a closed boundary. Note
that $(Lm(d_j),Fm(d_{j+1}))\sphat$ cannot be undefined for any $j\in
\{1,\dots,p-1\}$ (equation~\ref{eq:move_between_seg}). Moreover, one can
show that if the sequence defines a closed boundary and if
$Lm(d_p)=Fm(d_1)^{-1}$, then we should have $\alpha_i(d_p)= d_1$, which is
refused by hypothesis.

Using the same notations and hypothesis as equation~\ref{eq:or_seq},
one important result shown by Braquelaire and
Domenger~\cite{braquelaire-96-1,brun-02-5} states that the orientation of a
sequence $d_1\dots,d_p$ defining a closed boundary is equal to $4$ if
it is traversed clockwise and $-4$ otherwise. Moreover, this sequence
corresponds to:
\begin{itemize}
\item a finite face of $\dual{G_i}$ and thus a region if its
  orientation is equal to $-4$,
  
\item a set of faces of $\dual{G_i}$ connected by bridges and
contained in one face if the orientation is equal to $4$. Such a set
of faces is called an infinite face~\cite{braquelaire-96-1,brun-02-5}. It
encodes a connected component of the partition
(Section~\ref{sec:intro}).
\end{itemize}

By construction each combinatorial map $G_i$ of a combinatorial
pyramid is connected and all but one faces of $\dual{G_i}$ define a
finite face. The infinite face of a combinatorial map encodes the
background of the image (Section~\ref{sec:combi_maps}).


\section{Computing contains/inside relationships}
\label{sec:inclusion}

\begin{figure}[htbp]
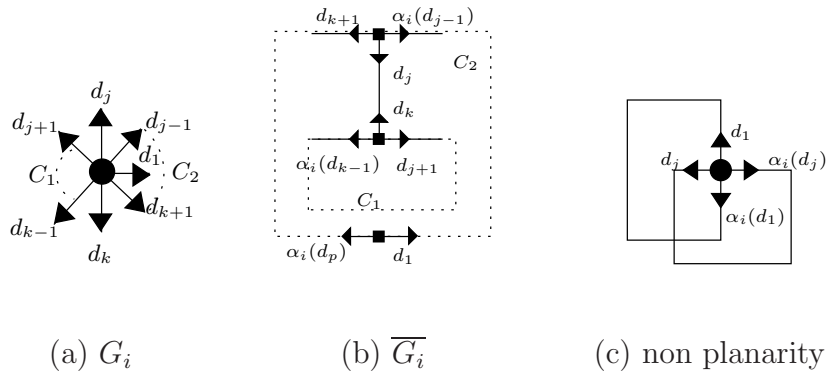

  \centering
  \begin{tabular}[c]{cp{5mm}cp{5mm}c}
    \input{proofg}&&\input{proof}&&\input{proof2}\\
    (a) $G_i$&&(b) $\dual{G_i}$&&(c) non planarity\\
  \end{tabular}

\caption{The local configuration in $G_i$ (a) and $\dual{G_i}$(b)  of the darts used by Propositions~\protect\ref{prop:rm_bridge} and~\ref{prop:or_C1}. Note that we implicitly suppose here that $d_j$ is the starting dart since $C_2$ surrounds $C_1$. (c) A contradiction obtained in the proof of Proposition~\ref{prop:or_C1}. }
  \label{fig:proofs}
\end{figure}

As demonstrated in Fig.~\ref{fig:combi_inside}, the determination of
the \emph{contains} and \emph{inside} relationships requires to
determine which vertices are surrounded by a self-loop incident to a
given vertex. This ambiguity in the location of the self-loop is
related to the fact that the two darts of a self-loop play a symmetric
role in the $\sigma$ cycle to which they belong
(Section~\ref{subsec:problem}). The determination of the
\emph{contains} and \emph{inside} relationships requires thus to
define a criterion in order to differentiate the two darts of a
self-loop. This criterion is provided by the following proposition
(Fig.~\ref{fig:proofs}(a)):

\begin{prop}
\label{prop:rm_bridge}
Consider a combinatorial map $G_i$ defined at level $i$ of a
combinatorial pyramid such that $G_i$ does not contain any redundant
edge. Let us additionally consider the darts around a vertex
$\sigma_i^*(d_1)=(d_1,\dots,d_j,\dots,d_k,\dots,d_p)$ of $G_i$ and a
self-loop $\alpha_i^*(d_j)=(d_j,d_k)$ such that dart $d_j$ is encountered
before $d_k$ when traversing $\sigma_i^*(d_1)$ from $d_1$, e.g. $j<k$. The
two sequences of darts $C_1=(d_{j+1},\dots,d_{k-1})$ and
$C_2=(d_{k+1},\dots,d_p,d_1,\dots,d_{j-1})$ define closed boundaries
and have an opposite orientation ($or(C_1)=-or(C_2)$). Moreover, the
two couple of darts $(d_{k-1},d_{j+1})$ and $(d_{k+1},d_{j-1})$ do not
define self-loops e.g. $d_{j+1}\neq\alpha_i(d_{k-1})$ and
$d_{j-1}\neq\alpha_i(d_{k+1})$.
\end{prop}
\begin{pf}
  First note that since $G_i$ does not contain empty self-loops both
  $C_1$ and $C_2$ should be non-empty.  

  Let us show that $C_1$ defines a closed boundary. The definitions of
  $\sigma_i^*(d_1)$ and $\alpha_i^*(d_j)$ induce the two following equalities:
  $\varphi_i(\alpha_i(d_{k-1}))=\sigma_i(d_{k-1})=d_k$ and
  $\varphi_i(d_k)=\sigma_i(d_j)=d_{j+1}$.  We have thus
  $d_{j+1}=\varphi_i^2(\alpha_i(d_{k-1}))$ which induces $d_{j+1}\in
  \varphi_i^*(\alpha_i(d_{k-1}))$. The same arguments are used to show that $C_2$
  defines a closed boundary.

  Let us now show that $d_{j+1}\neq\alpha_i(d_{k-1})$. Since
  $d_{j+1}=\varphi_i^2(\alpha_i(d_{k-1}))$, the relation $d_{j+1}=\alpha_i(d_{k-1})$
  implies that $d_{j+1}=\varphi_i^2(d_{j+1})$. The dart $d_{j+1}$ would thus
  be incident to a degree two face which is refused by hypothesis
  since $G_i$ does not contain empty double edges. The same argument
  is used to show that $d_{j-1}\neq\alpha_i(d_{k+1})$.

  All the conditions to apply equation~\ref{eq:or_seq} are thus
  satisfied and we derive:
\begin{equation}
\label{eq:rm_bridge}
or(\sigma_i^*(d))=or(C_1)+or(C_2)-4
\end{equation}
where $or(\sigma_i^*(d))$ denotes the orientation of the whole sequence of
darts $(d_1,\dots,d_p)$. Since this sequence defines a
counter-clockwise traversal of the face its orientation is equal to
$-4$ (Section~\ref{subsec:seg_or}). We have thus $or(C_1)=-or(C_2)$.
$\Box$
\end{pf}

Proposition~\ref{prop:rm_bridge} may be interpreted as follows: The loop
$\alpha_i^*(d_j)$ corresponds to a bridge in $\dual{G_i}$ the removal of
which splits the combinatorial map into two connected components. The
component encoding the surrounding face is traversed counter-clockwise
and has thus an orientation of $-4$. The remaining component
corresponds to the connected component of inside regions and has an
opposed orientation of $4$ (section~\ref{subsec:seg_or}). We say that
$d_j$ is the starting dart of the loop if the sequence of darts
encoding the inside connected component is enclosed between $d_j$ and
$d_k=\alpha_i(d_j)$. This property is thus characterized by
$or(C_1)=4$. The dart $d_j$ is called the ending dart of the loop
otherwise. Note that if $d_j$ is a starting dart $\alpha_i(d_j)$ should be
an ending dart and conversely.

The above discussion and Proposition~\ref{prop:rm_bridge} provide thus a
criterion which differentiates the two darts of a loop in order to
characterize the inside relationship. However, computing the
orientation of all sequences of darts between the two darts of all
self-loops incident to a vertex would require extra calculus. Indeed,
nested self-loops may induce several traversals of a same sequence of
darts. The following theorem incrementally computes the orientation of
any sequence of darts surrounded by the two darts of a loop:
\begin{prop}
  \label{prop:or_C1}
  Using the same hypothesis and notations as
  Proposition~\ref{prop:rm_bridge}, the orientation of the sequence of
  darts $C_1=d_{j+1}\dots.d_{k-1}$ between $d_j$ and $d_k$ is defined
  by :
  \[
  or(C_1)=or'(d_1\dots.d_{k-1})-or'(d_1\dots.d_j)-(Lm(d_j),Fm(d_{j+1}))\sphat+(Lm(d_{k-1}),Fm(d_{j+1}))\sphat
  \]
where $or'(d_1\dots.d_{k-1})$ and $or'(d_1\dots.d_j)$ are the
orientations of the sequences $d_1\dots.d_{k-1}$ and $d_1\dots.d_j$
(equation~\ref{eq:or_seq}) considered as non-closed sequences of
darts.
\end{prop}
\begin{pf}  
We want to use equation (12) to calculate the orientation of the
  sequences $C_1 = (d_{j+1},\dots,d_{k-1})$, $(d_1,\dots,d_j)$ and
  $(d_1,\dots,d_{k-1})$.
\begin{enumerate}
\item $C_1$: The precondition $d_{k-1} \not= \alpha_i(d_{j+1})$ is
  equivalent to $d_{j+1} \not= \alpha_i(d_{k-1})$ which is excluded in
  Proposition~\ref{prop:rm_bridge}. Moreover, $C_1$ is a closed
  sequence.

\item $(d_1,\dots,d_j)$: If $d_1=\alpha_i(d_j)$ then $d_1=d_k$ (since
$\alpha_i(d_j)=d_k$) and thus $k=1$ and $p=k-1$. This last result
contradicts our hypothesis $j<k$.
\item $(d_1,\dots,d_{k-1})$ : If we assume that $d_{k-1} = \alpha_i(d_1)$
and combine it with the relationship
$\sigma_i(d_{k-1})=d_k=\alpha_i(d_j)$ we can express $\sigma_i^*(d_1)$ as:
\begin{equation}
  \begin{array}[c]{lllccl}
    \sigma_i^*(d_1)&=&(d_1,\dots,d_j,\dots,&d_{k-1},&d_k&,\dots,d_p)\\
    &=&(d_1,\dots,d_j,\dots,&\alpha_i(d_1),&\alpha_i(d_j)&,\dots,d_p)\\
\end{array}
\label{eq:planarity}
\end{equation}
This last equation contradicts the planarity of $G_i$ since the edges
$\alpha_i^*(d_1)$ and $\alpha_i^*(d_j)$ must cross in order to satisfy
equation~\ref{eq:planarity} (Fig.~\ref{fig:proofs}(b)).
\end{enumerate}
We now can expand the orientations of the three sequences involved in
Proposition~\ref{prop:or_C1} to show that
$or(C1)-or'(d_1,\dots,d_k)+or'(d_1,\dots,d_j) =
(Lm(d_{k-1}),Fm(d_{j+1}))\sphat-(Lm(d_j),Fm(d_{j+1}))\sphat$. Indeed
$or(C1)-or'(d_1,\dots,d_k)+or'(d_1,\dots,d_j)$ may be expanded as
follows:
\[
\begin{array}[c]{lp{6mm}c}
\sum\limits_{r=j+1}^{k-2}\left(or(d_r)+(Lm(d_r),Fm(d_{r+1}))\sphat\right)\\
+or(d_{k-1})+(Lm(d_{k-1}),Fm(d_{j+1}))\sphat&& (or(C_1))\\
-\left(\sum\limits_{r=1}^{k-2}\left(or(d_r)+(Lm(d_r),Fm(d_{r+1}))\sphat\right)+
or(d_{k-1})\right)&& (or'(d_1,\dots,d_{k-1}))\\
+\left(\sum\limits_{r=1}^{j}or(d_r)+(Lm(d_r),Fm(d_{r+1}))\sphat-(Lm(d_j),Fm(d_{j+1}))\sphat\right)&& \left(or'(d_1,\dots,d_{j})\right)\\
=(Lm(d_{k-1}),Fm(d_{j+1}))\sphat-(Lm(d_j),Fm(d_{j+1}))\sphat~. \Box\\
\end{array}
\]

\end{pf}

  \begin{algorithm}
    \begin{center}
      \begin{minipage}{14cm}
        \small\tt 
        \begin{tabbing}xxxx\=xxxx\=xxxx\=xxxx\=xxxx\=xxxx\=xxxx\=xxxxxxx\=xxxx\=xxxx\=xxxx\=\kill
               1\>list starting\_dart(combi\_map $G_i$, dart $d_1$) \{\\
     2\>\>list L=$\emptyset$\\
     3\>\>stack P\\
     4\>\>for each dart $d_k$ in $\sigma_i^*(d)=(d_1,\dots,d_p)$\{\\
     5\>\>\>if($d_k$ is a loop) \{\\
     6\>\>\>\>if(P is empty or $\alpha_i(d_k)$ is not on the top of the stack P)\\
     7\>\>\>\>\>push $d_k$ and $or(d_1,\dots,d_k)$ in P\\
     8\>\>\>\>else \{// $\alpha_i(d_k)$ on top of the stack P\>\\
     9\>\>\>\>\>let $C_1$ be the sequence of darts between $\alpha_i(d_k)$ and $d_k$\\
    10\>\>\>\>\>computes $or(C_1)$ using Proposition~\ref{prop:or_C1}\\
    11\>\>\>\>\>if($or(C_1)==4$) $L=L\cup\{\alpha_i(d_k)\}$ else $L=L\cup\{d_k\}$ \>\\
    12\>\>\>\>\}\\
    13\>\>\}\\
    14\>\>return L\\
    15\>\}\\

        \end{tabbing}
      \end{minipage}
      \caption{\protect\small\it Determination of the starting darts of the  loops}
      \label{alg:open_parenthesis2}
    \end{center} 
  \end{algorithm}
   Propositions~\ref{prop:rm_bridge} and~\ref{prop:or_C1} are
the basis of the algorithm {\tt staring\_darts}
(Algorithm~\ref{alg:open_parenthesis2}) which traverses the $\sigma_i$
cycle of a given vertex $\sigma_i^*(d_1)=(d_1,\dots,d_p)$ and computes at
each step the orientation of the sequence $d_1\dots,d_k$. Using the
same notations as Proposition~\ref{prop:rm_bridge}, let us consider a loop
$\alpha_i^*(d_j)=(d_j,d_k)$ such that $d_j$ has been previously encountered
by the algorithm ($j<k$). The algorithm \texttt{starting\_dart}
determines the starting dart between $d_j$ and $d_k$ on lines 10 and
11 from the orientation of $C_1=(d_{j+1}\dots,d_{k-1})$ by using
Propositions~\ref{prop:rm_bridge} and~\ref{prop:or_C1}.  This
starting dart is added to a list returned by the algorithm.

Since the loops are nested $d_j$ and $or'(d_1.\dots.d_j)$ should be on
the top of stack $P$ used by the algorithm. The darts $d_{j},d_{j+1}$
and $d_{k-1}$ are retrieved from the current dart $d_k$ by: \(
d_j=\alpha_i(d_k)\;;\;d_{j+1}=\sigma_i(d_j)\;\mbox{ and }\;d_{k-1}=\sigma_i^{-1}(d_k)
\). Moreover, the orientation of $C_1$ (Proposition~\ref{prop:or_C1})
is evaluated in constant time since $or'(d_1,\dots,d_{k-1})$ is the
last orientation and $or'(d_1\dots,d_j)$ is retrieved from the stack.

Given the list of starting darts determined by the algorithm {\tt
stating\_darts}, the set of vertices contained in $\sigma_i^*(d_1)$ is
retrieved by traversing, the sequence $\sigma_i^*(d_1)$ from each starting
dart to the corresponding end. By construction all darts encountered
between the starting and ending darts of the loop encode adjacency
relationships to contained vertices.  Note that in case of nested
loops some loops may be traversed several times. Given a starting dart
$d$, this last drawback may be avoided by replacing any encountered
starting dart by its $\alpha_i$ successor during the traversal from $d$ to
$\alpha_i(d)$.

Our algorithm, is thus local to each vertex and the method may be
applied in parallel to all the vertices of the combinatorial map
$G_i$. Given a vertex $\sigma_i^*(d_1)$, the determination of its starting
darts (algorithm {\tt starting\_darts}) requires to traverse once
$\sigma_i^*(d_1)$. Moreover, the determination of the inside relationships
from the list of starting darts requires to traverse each dart of
$\sigma_i^*(d_1)$ at most once. The worse complexity of our algorithm is thus
bounded by the maximum degree of a vertex,
e.g. $\complex{2|\sigma_i^*(d_1)|}$.

\subsection{Application to road sign's recognition}
\label{subsec:application}


    
\begin{figure}[t]
  \centering
  \begin{tabular}[b]{cccc}
    \epsfig{file=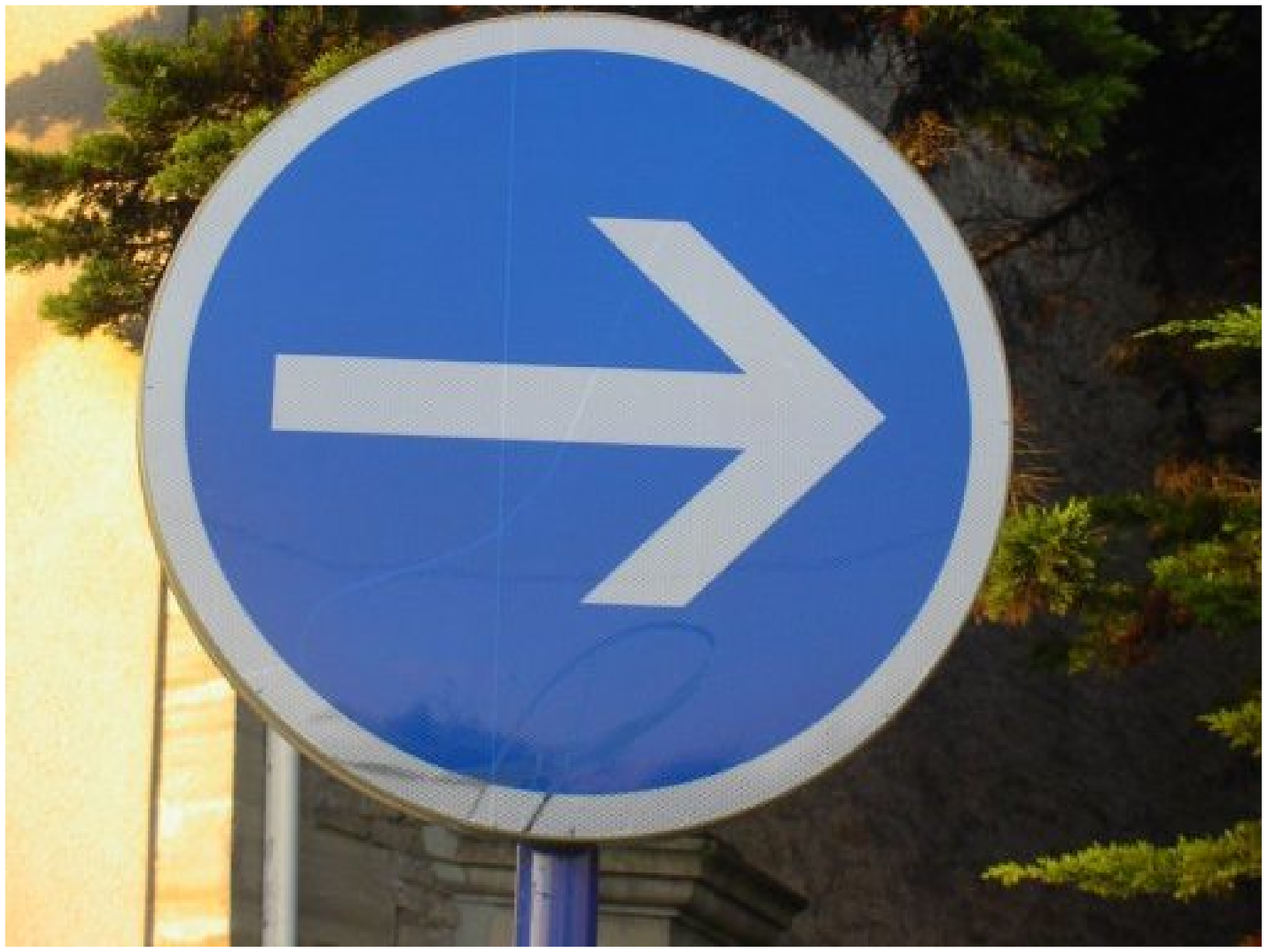,height=1.5cm}&
    \epsfig{file=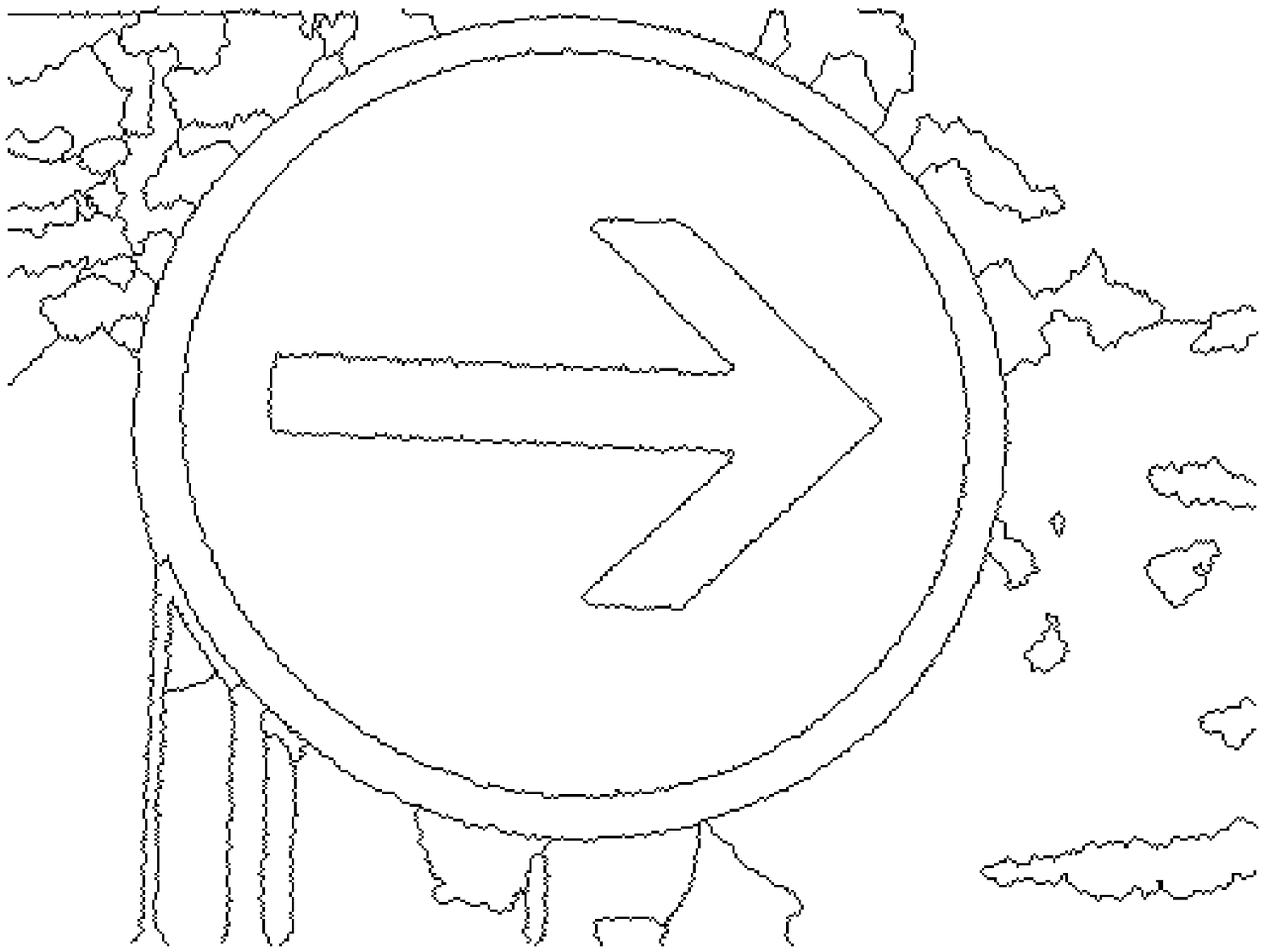,height=1.5cm}&
    \epsfig{file=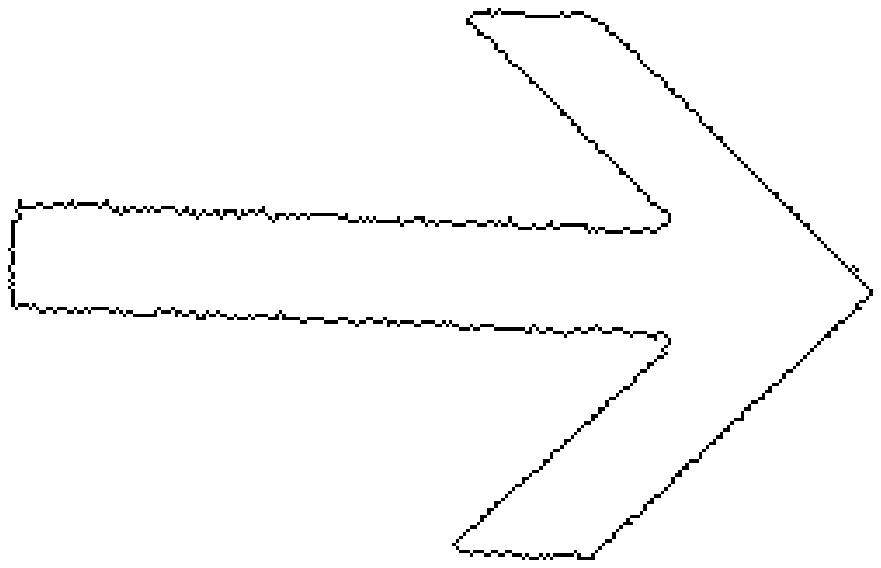,height=1.5cm}&
    {  \unitlength= .4mm
  \begin{picture}(50,50)
    \put(25,25){\color{blue}\circle*{5}}
    \put(10,25){\circle{5}}
    \put(40,25){\circle{5}}
    \put(20,25){\line(-1,0){5}}
    \put(30,25){\line(1,0){5}}

    \put(25,29.5){\line(0,1){8}}
    \put(25,37.5){\line(-1,0){20}}
    \put(5,37.5){\line(0,-1){30}}
    \put(5,7.5){\line(1,0){20}}
    \put(25,7.5){\line(0,1){12}}

  \end{picture}}\\
    \epsfig{file=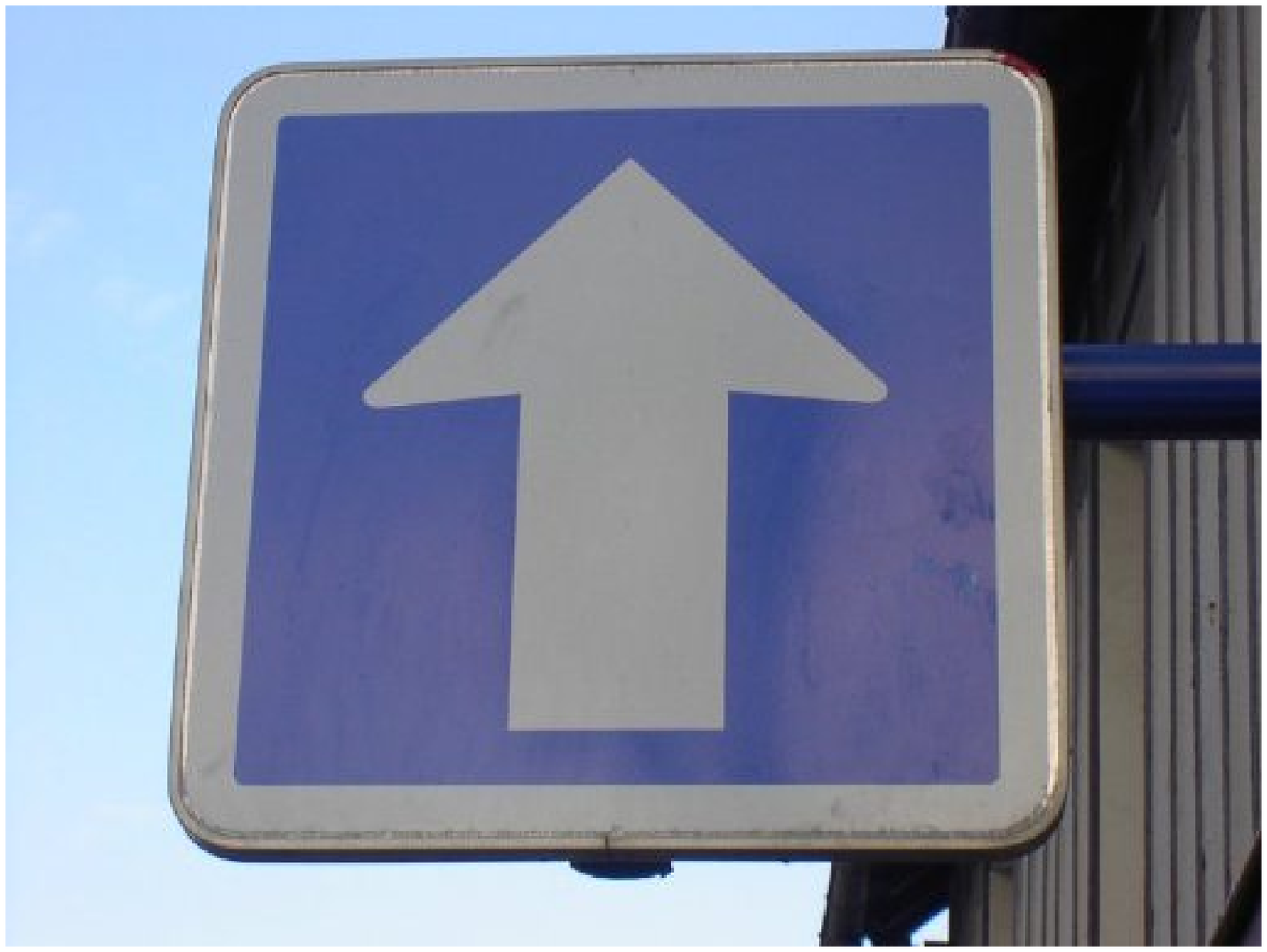,height=1.5cm}&
    \epsfig{file=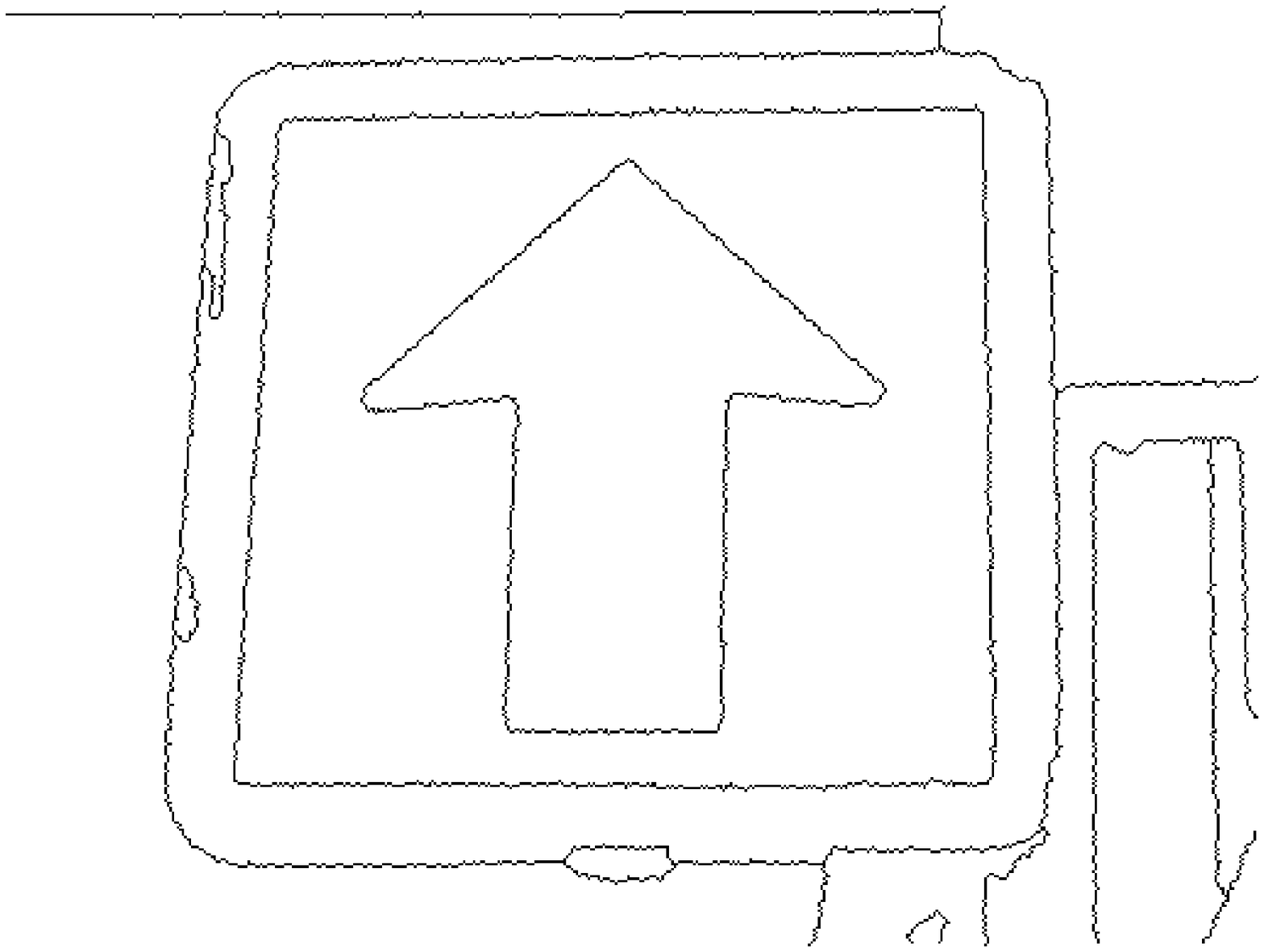,height=1.5cm}&
    \epsfig{file=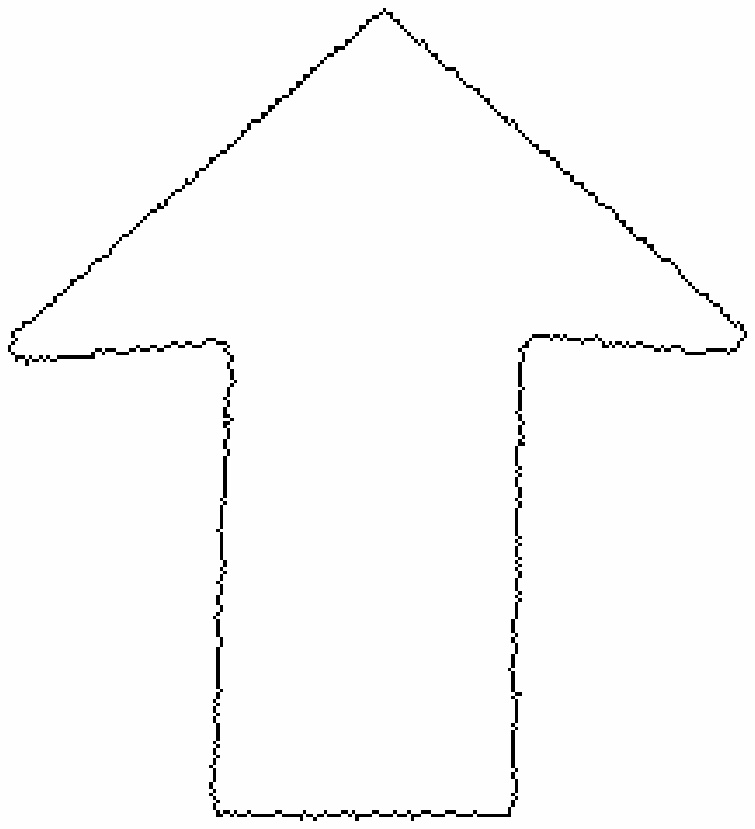,height=1.5cm}\\
    (a)&(b)&(c)&(d)\\
    
  \end{tabular}
  \caption{Extraction of  symbols within roadsigns using contains/inside information.}
  \label{fig:roadsign}
\end{figure}

Fig.~\ref{fig:roadsign} illustrates one application of the
contains/inside information to image analysis. The road sign shown in
Fig.~\ref{fig:roadsign}(a) is composed of only two colors with one
symbol inside a uniform background, the background itself being
surrounded by one border with a same color as the symbol. In our
example, the two roadsigns have a uniform background which includes
one symbol representing a white arrow. The background is surrounded by
a white border. In this application we wish to extract the sign using
only topological and color information (and thus independently of the
shapes of the symbol and the road sign). Using only adjacency and
color information, the symbol cannot be distinguished from the border
of the road sign since the border and the symbol have a same color and
are both adjacent to the background of the road sign
(Fig.~\ref{fig:roadsign}(d)). However, using contains/inside
information, the symbol and the border may be distinguished since the
border contains the background of the road sign which contains the
symbol. Our algorithm first builds a combinatorial pyramid using a
hierarchical watershed
algorithm~\cite{brun-05}. Fig.~\ref{fig:roadsign}(b) shows the top
level of the hierarchies obtained from the two roadsigns.  Using the
top level combinatorial map of each pyramid our algorithm selects the
$k$ regions of the partition whose color is closest from the
background's color ($k$ is fixed to five in our experiment). This last
step defines a set of candidate regions for the background of the road
sign.  This background is then determined as the region whose
contained regions have the closest mean color from the color's symbol
(equal to white in this experiment). Note that this step removes from
the $k$ selected candidates any regions which do not contain another
region. We thus make explicit the a priori knowledge that the
background of the road sign should contain at least one region. The
symbol is then determined as the set of regions inside the selected
region (Fig.~\ref{fig:roadsign}(c)).  Finally, let us note that the
contains information needs to be computed only on the $k$ selected
candidates for the road sign's background. Within this experiment a
global algorithm computing the contains information for all vertices
would require useless calculus.


\section{Conclusion}

We have introduced in this paper 5 relationships between regions
(Section~\ref{subsec:relating}).  These relationships are devoted to
the graph based segmentation framework and encode either rough or fine
relationships between the regions of a partition: The
\emph{meets\_exists} relationship corresponds to the ability of a model
to encode the existence of at least one common boundary between two
regions.  The \emph{meets\_each} relationship corresponds to an
encoding of each connected boundary between two adjacent regions. The
\emph{inside} and \emph{contains} relationships are asymmetric and
encode the fact that one region contains the other. Finally, the
\emph{composed of} relationships is only provided by hierarchical data
structures and encodes the fact that one region is composed of several
regions defined at levels below.

Table~\ref{tab:relations} shows the ability of the Region Adjacency
Graph, the combinatorial map, the simple graph pyramid, the dual graph
pyramid and the combinatorial pyramid to encode the
\emph{meets\_exist}, \emph{meets\_each}, \emph{contains/inside} and
\emph{composed of} relationships. The Region Adjacency Graph
(Section~\ref{subsec:rag}) only encodes the \emph{meets\_each}
relationship. The combinatorial map model
(Section~\ref{subsec:combi_maps}) encodes all but the \emph{composed
of } relationships. The simple graph pyramids
(Section~\ref{sec:irr_pyr}) encodes the \emph{meets\_exist} and the
\emph{composed of } relationships. This last relationship is also
encoded by the two other irregular pyramid models described in this
paper (section~\ref{subsec:irr_pyr}). The dual graph pyramids
(Section~\ref{subsec:dual_graph}) encodes the \emph{meets\_exists},
\emph{meets\_each} (Section~\ref{subsec:dual_graph_meet_each}) and
\emph{composed of} relationships. The \emph{inside/contains}
relationships can not be deduced from the model using local calculus
(Section~\ref{subsec:dual_graph_inside}). However, the authors
conjecture that these relationships may be computed using the fact
that the vertex encoding the background of the image is not surrounded
by any self-loop. Such an algorithm would require a propagation step
from the background vertex and would thus require global
calculus. This property is indicated by an interrogation mark in
Table~\ref{tab:relations}. The combinatorial map pyramid model
(Section~\ref{sec:combi_pyr}) encodes the \emph{meets\_exists},
\emph{meets\_each} (Section~\ref{subsec:problem}) and \emph{composed
of} relationships.

The main contribution of this paper consists in the design of the
algorithm {\tt starting\_dart} (Section~\ref{sec:inclusion}) which uses
the orientation explicitly encoded by combinatorial maps to
differentiate the two darts of a self-loop.  Given a vertex incident
to a self-loop, this last characterization allows to determine the
regions \emph{inside} the region encoded by this vertex in a time
proportional to twice its number of incident edges.  This method
implies only local calculus and its parallel complexity is bounded by
twice the maximal degree of the vertices of the graph.

The efficient computation of those relations relating regions of a
segmentation is a prerequisite to the description and the recognition
of relevant groupings: an important step on the way to more generic
recognition, categorization and higher visual abstraction within the
homogeneous framework of combinatorial pyramids.

\bibliographystyle{elsart-num}
\bibliography{comp-graph}

\begin{thebibliography}{10}
\expandafter\ifx\csname url\endcsname\relax
  \def\url#1{\texttt{#1}}\fi
\expandafter\ifx\csname urlprefix\endcsname\relax\def\urlprefix{URL }\fi

\bibitem{jiang-01}
X.~Jiang, A.~Munger, H.~Bunke, On median graphs: properties, algorithms and
  applications, IEEE Transactions on Pattern Analysis and Machine Intelligence
  23~(10) (2001) 1114--1151.

\bibitem{randell-92}
D.~Randell, C.~Z, A.~Cohn, A spacial logic based on regions and connections,
  in: B.~Nebel, W.~Swartout, C.~Rich (Eds.), Principle of Knowledge
  Representation and Reasoning: Proceedings 3rd International Conference,
  Cambridge MA, 1992, pp. 165--176.

\bibitem{shearer-01}
K.~Shearer, H.~Bunke, S.~Venkatesh, Video indexing and similarity retrieval by
  largest common subgraph detection using decision trees, Pattern Recognition
  34 (2001) 1075--1091.

\bibitem{tweed-03}
D.~S. Tweed, A.~D. Calway, Integrated segmentation next term and depth ordering
  of motion layers in image sequences, Image and Vision Computing 20~(9) (2003)
  709--723.

\bibitem{braquelaire-96-1}
J.~P. Braquelaire, L.~Brun, Image segmentation with topological maps and
  inter-pixel representation, Journal of Visual Communication and Image
  representation 9~(1) (1998) 62--79.

\bibitem{fiorio-95}
C.~Fiorio, Approche interpixel en analyse d'images~: une topologie et des
  algorithmes de segmentation, Th{\`e}se de doctorat, Universit{\'e}
  Montpellier II (24 novembre 1995).

\bibitem{brun-02-5}
L.~Brun, M.~Mokhtari, J.~P. Domenger, Incremental modifications on segmented
  image defined by discrete maps, Journal of Visual Communication and Image
  Representation 14 (2003) 251--290.

\bibitem{montanvert-91}
A.~{M}ontanvert, P.~Meer, A.~Rosenfeld, Hierarchical image analysis using
  irregular tessellations, IEEE Transactions on Pattern Analysis and Machine
  Intelligence 13~(4) (1991) 307--316.

\bibitem{meer-89}
P.~Meer, Stochastic image pyramids, Computer Vision Graphics Image Processing
  45 (1989) 269--294.

\bibitem{brun-02-9}
L.~Brun, Traitement d'images couleur et pyramides combinatoires, Habilitation
  {\`a} diriger des recherches, Universit{\'e} de Reims (2002).

\bibitem{willersinn-94}
D.~Willersinn, W.~G. Kropatsch, Dual graph contraction for irregular pyramids,
  in: International Conference on Pattern Recogntion D: Parallel Computing,
  International Association for Pattern Recognition, Jerusalem, Israel, 1994,
  pp. 251--256.

\bibitem{braquelaire-01}
J.~P. Braquelaire, P.~Desbarats, J.~P. Domenger, 3d split and merge with
  3-maps, in: J.~M. {J}olion, W.~Kropatsch, M.~Vento (Eds.), $3^{rd}$ Workshop
  on Graph-based Representations in Pattern Recognition, IAPR-TC15, CUEN,
  Ischia(Italy), 2001, pp. 32--43.

\bibitem{bertrand-01}
Y.~Bertrand, G.~Damiand, C.~Fiorio, Topological map: Minimal encoding of 3d
  segmented images, in: J.~M. {J}olion, W.~Kropatsch, M.~Vento (Eds.), $3^{rd}$
  Workshop on Graph-based Representations in Pattern Recognition, IAPR-TC15,
  CUEN, Ischia(Italy), 2001, pp. 64--73.

\bibitem{brun-03-1}
L.~Brun, W.~Kropatsch, Combinatorial pyramids, in: Suvisoft (Ed.), IEEE
  International conference on Image Processing (ICIP), Vol.~II, IEEE,
  Barcelona, 2003, pp. 33--37.

\bibitem{brun-05}
L.~Brun, M.~Mokhtari, F.~Meyer, Hierarchical watersheds within the
  combinatorial pyramid framework, in: Proc. of DGCI 2005, Vol. 3429,
  IAPR-TC18, LNCS, 2005, pp. 34--44.

\bibitem{damiand-05}
G.~Damiand, M.~Dexet-Guiard, P.~Lienhardt, E.~Andres, Removal and contraction
  operations to define combinatorial pyramids: application to the design of a
  spatial modeler, Image and Vision Computing 23~(2) (2005) 259--269.

\bibitem{Kropatsch95b}
W.~G. {{K}ropatsch}, Building {I}rregular {P}yramids by {D}ual {G}raph
  {C}ontraction, IEE-Proc. Vision, Image and Signal Processing Vol.~142~(No.~6)
  (1995) pp.~366--374.

\bibitem{Fermueller94a}
C.~{Ferm{\"u}ller}, W.~G. {Kropatsch}, A {S}yntactic {A}pproach to
  {S}cale-{S}pace-{B}ased {C}orner {D}escription, IEEE Transactions on Pattern
  Analysis and Machine Intelligence Vol.~16~(No.~7) (1994) pp.~748--751.

\end{thebibliography}




\input{tables}
\end{document}